\documentclass[runningheads]{llncs}
\usepackage{graphicx}
\begin{document}
\title{Towards Using Multiple Iterated, Reproduced, and Replicated Experiments with Robots (MIRRER) for Evaluation and Benchmarking\thanks{This work was supported in part by the National Science Foundation (awards CNS-1925604 and TI-2229577) and the National Institute of Standards and Technology (award 70NANB20H199).}}




%
\titlerunning{MIRRER Evaluation and Benchmarking Framework}
%
\author{Adam Norton\inst{1}\orcidID{0000-0002-6127-4588} \and
Brian Flynn\inst{1}\orcidID{0000-0001-5549-4528}}
\authorrunning{Norton and Flynn}
%
\institute{New England Robotics Validation and Experimentation (NERVE) Center,\\University of Massachusetts Lowell, Lowell, MA, 01852, USA
\email{adam\_norton,brian\_flynn@uml.edu}\\
\url{https://nerve.uml.edu/}}
\maketitle              
\begin{abstract}
The robotics research field lacks formalized definitions and frameworks for evaluating advanced capabilities including generalizability (the ability for robots to perform tasks under varied contexts) and reproducibility (the performance of a reproduced robot capability in different labs under the same experimental conditions).
This paper presents an initial conceptual framework, MIRRER, that unites the concepts of performance evaluation, benchmarking, and reproduced/replicated experimentation in order to facilitate comparable robotics research.
Several open issues with the application of the framework are also presented.
\end{abstract}
\section{Introduction and Background}

The benefit of advanced robotics over traditional automation is the ability for robotic capabilities to generalize across domains and applications; in other words, a robotic component or system has the ability to perform a task with some level of variation. 
This capability is referred to by many names including robustness, flexibility, versatility, and generalizability, all of which refer to the robot’s ability to operate when variations are induced.
Open issues with developing generalizable robotics solutions include a lack of accepted frameworks and terminology to properly define their capabilities as well as a lack of effective benchmarking methodologies. 
The use of learning-based algorithms in robotics allows for generalizable solutions to be developed; however, issues with reproducibility and replicability -- of the experimental conditions and, more importantly, of the results -- are incurred.
The non-deterministic nature of these approaches coupled with the complexity of a physical robot system operating in a real-world environment (although similar issues  exist in simulation) is problematic.

While issues with reproducibility and replicability are not unique to robotics research, there is a multitude of evidence in the field that they are prominent (e.g., issues with reusing code~\cite{cervera2018try,cervera2023run}, reproducing deep reinforcement learning~\cite{lynnerup2020survey}, multiple human-robot interaction specific concerns~\cite{gunes2022reproducibility,leichtmann2022crisis}) with some efforts towards solving them (e.g., IEEE R-articles~\cite{bonsignorio2015toward,bonsignorio2017new}, ACM badging to denote reproduced or replicated results~\cite{acmbadges}).
Particularly in the domain of robotic manipulation, benchmarking tools have also been developed to bolster the repeatability of physical robot testing (e.g., accurately repositioning artifacts via automated mechanisms~\cite{burgess2022dgbench,dufrene2024icra} or augmented reality~\cite{khargonkar2023scenereplica}) and the reproducibility of research results (e.g., task protocols to benchmark performance~\cite{leitner2017acrv,bekiroglu2019benchmarking}, datasets for comparing results~\cite{morrison2020egad}).
A survey conducted of the robotics research community and a series of workshops on the current state of open-source assets and benchmarking resources reveals that a lack of consensus on metrics, protocols, software component structures, and incentives are among the open issues impacting reproducibility and replicability~\cite{posecompare}.

If a robotic capability is purported to be generalizable, one should be able to evaluate this assertion by iterating on the conditions of an experiment and determining to what level the two varied conditions impact performance.
Similarly, the reproducibility of a robotic capability should also be able to be evaluated by reproducing the conditions of an experiment (or set of experiments), which can then also be used to evaluate generalizability.
Lastly, the same techniques used to reproduce an experiment should also allow for different but comparable robotic capabilities to be evaluated and compared across labs (referred to as replicability, defined later).
We propose to unite these concepts under a single framework in order to formalize their definitions, articulate their interrelated nature when it comes to conducting evaluations, and justify the development of new methodologies and software/hardware architectures to improve robot benchmarking.
While all issues are not addressed in this paper, an initial conceptual framework (MIRRER) is presented that unites these concepts towards solving the challenges associated with evaluating, reproducing, and replicating robotics research.



\section{MIRRER Framework}

Throughout this section, all examples provided use a robotic grasping experiment for a pick-and-place task as a notional scenario.
We first provide proposed definitions for the concepts and terminology used in the Multiple Iterated, Reproduced, and Replicated Experiments with Robots (MIRRER) framework: 
\begin{itemize}
    \item \textbf{Context} of a robotic experiment refers to the  parameters of the system's operation that can impact performance, distilled into four categories (adapted from \cite{norton2020developing}):
    \begin{itemize}
        \item \textbf{Input data} provided to the robot to perform its task (e.g., 3D models of objects with pre-planned grasps, point clouds collected in-situ).
        \item \textbf{Target objects} being interacted with (e.g., objects to be grasped, kit form obstructions to avoid colliding with).
        \item \textbf{Tasks} being performed with or around those objects (e.g., grasp object, pick up, and place in kit).
        \item \textbf{Environment} where the task is being performed (e.g., ambient lighting, layout of bin with objects and kits).
    \end{itemize}
    \item \textbf{Robot system} executing the tasks whose configuration consists of multiple components including perception modules, motion planners, grasp planners, hardware, etc. The component that is being evaluated through experimentation is referred as the \textbf{component under evaluation (CUE)}.
    \item \textbf{Generalizability} of the CUE's performance is a metric that is evaluated when multiple experiments are conducted in the same or different labs under intentionally varied contexts (e.g., light vs. dark lighting conditions). 
    \item \textbf{Iterated} experiments are conducted in the same lab and use either:
    \begin{itemize}
        \item Different context parameters and the same robot system configuration to evaluate \textbf{generalizability} of the CUE's performance (e.g., evaluating a grasp planner using two sets of target objects), or
        \item The same context parameters and a different robot system configuration to \textbf{compare} the performance of multiple CUEs (e.g., comparing the performance of two grasp planners).
    \end{itemize}
    \item \textbf{Reproduced} experiments are conducted in a different lab, using the same context parameters and the same robot system configuration to evaluate \textbf{reproducibility} of the CUE's performance (e.g., lab 2 reproduces an experiment conducted by lab 1 to evaluate the performance of a grasp planner); adapted from \cite{acmbadges,joint2012international}. 
    \item \textbf{Replicated} experiments are conducted in a different lab, using the same context parameters and a different robot system configuration to \textbf{compare} the performance of multiple CUEs (e.g., comparing the performance of two grasp planners); adapted from \cite{acmbadges,joint2012international}.
\end{itemize}

Another commonly used term is ``repeat''; while the proposed framework does not use this term explicitly, experiments that are conducted in the same lab, using the same context parameters and with the same robot system configuration (e.g., conducting multiple grasping trials on the same object to achieve statistically significant results) are \textbf{repeated} experiments (adapted from \cite{acmbadges,joint2012international}).
It should be noted that ``reproducibility'' and ``replicability'' are sometimes swapped or used interchangeably in the research literature, but the proposed conceptual framework explicitly differentiates them, following suit with the National Information Standards Organization (NISO)~\cite{acmbadges} and the Joint Committee for Guides in Metrology (JCGM)~\cite{joint2012international}.

\begin{figure*}[b!]
\includegraphics[width=\textwidth]{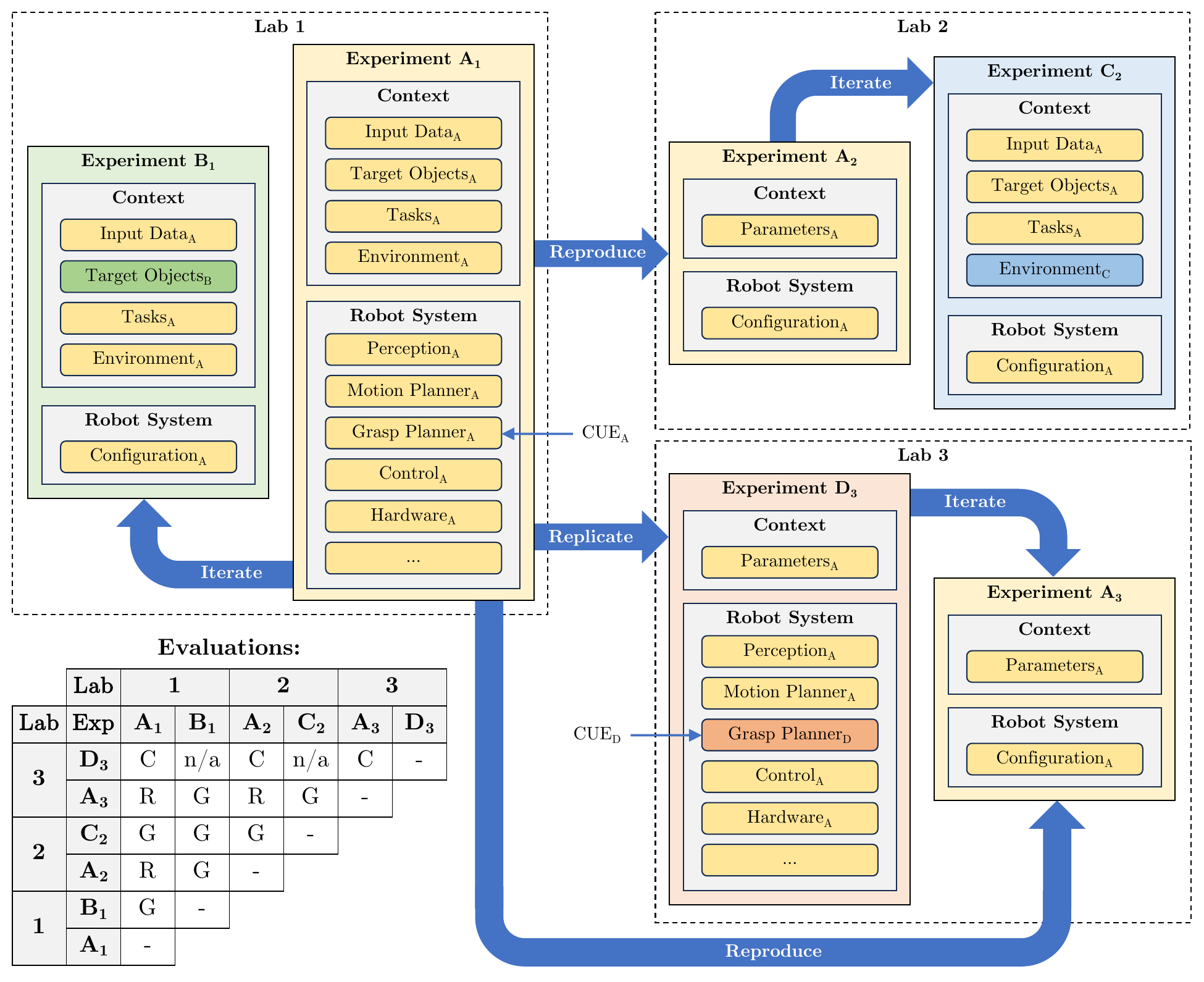}
\caption{The MIRRER framework using an example scenario involving six experiments across three labs to evaluate the performance of two CUEs (grasp planners). The table in the bottom left shows the evaluations that can be performed across experiments (G = generalizability, R = reproducibility, C = comparison, n/a = not applicable).} \label{fig1}
\end{figure*}

A diagram of the framework that relates the concepts of context, generalizability, reproducibility, comparison, and iterated, reproduced, and replicated experiments using an example scenario can be seen in Fig.~\ref{fig1}.
In this example representation of the framework, three labs each conduct two experiments: 
\begin{itemize}
    \item Lab 1 conducts experiment A\textsubscript{1} and \textbf{iterates} to conduct experiment B\textsubscript{1} with different target objects
    \item Lab 2 \textbf{reproduces} experiment A\textsubscript{1} to conduct experiment A\textsubscript{2} and \textbf{iterates} to conduct experiment C\textsubscript{2} with different lighting conditions
    \item Lab 3 \textbf{replicates} experiment A\textsubscript{1} to conduct experiment D\textsubscript{3} with a different grasp planner (CUE\textsubscript{D}) and \textbf{iterates} to conduct experiment A\textsubscript{3} using the original grasp planner (CUE\textsubscript{A}) simultaneously \textbf{reproducing} experiment A\textsubscript{1}
    \item Across these six experiments, the following evaluations can be conducted:
    \begin{itemize}
        \item \textbf{Generalizability} of CUE\textsubscript{A} can be evaluated 7 times
        \item \textbf{Reproducibility} of CUE\textsubscript{A} can be evaluated 3 times
        \item \textbf{Comparing} the performance of CUE\textsubscript{A} and CUE\textsubscript{D} 3 times
    \end{itemize}
    
\end{itemize}

While a reproduced experiment may use the same context parameters and robot system configuration, there will be natural variations due to the complexities of real-world environments and nuances of physical systems.
These natural variations have been demonstrated to substantially impact reproduced experiment results; for example, a 20\% variation in performance was observed in \cite{dasari2022rb2} when reproducing an experiment across two labs.
The authors suggest that ``building precisely reproducible robotic setups is impossible and therefore absolute performance numbers on a benchmark task are meaningless.'' 
They instead recommend using a \textbf{local relative ranking (LRR)} evaluation method where each lab establishes a baseline within their own lab for comparison when benchmarking the performance of a new capability~\cite{dasari2022rb2}. 
So then, following this recommendation when benchmarking to compare the performance of different CUEs, one should conduct both a \textbf{reproduced} experiment (to produce a local baseline with the original CUE) and a \textbf{replicated} experiment (to produce performance results of the new CUE).
For example, using the scenario in Fig.~\ref{fig1}, A\textsubscript{3} produces a local baseline of CUE\textsubscript{A} in lab 3 that can then be compared to the results of D\textsubscript{3} which uses CUE\textsubscript{D} to generate LRR\textsubscript{3}.
It should be noted that the LRR method was developed with the evaluation of software components in mind; producing a local baseline when the CUE is a piece of hardware such as a custom gripper (e.g., Yale OpenHand~\cite{ma2017yale}) will introduce additional challenges (e.g., costs to purchase, fabrication expertise, logistics to share hardware between labs).

The authors of \cite{dasari2022rb2} also propose a method to globally rank several LRRs contributed by different labs.
For example, if another lab were added to the scenario in Fig.~\ref{fig1} (lab 4) and reproduced experiments A\textsubscript{3} and D\textsubscript{3} from lab (i.e., A\textsubscript{4} and D\textsubscript{4}), then LRR\textsubscript{3} and LRR\textsubscript{4} can be contributed towards a global ranking of the performance of CUE\textsubscript{A} and CUE\textsubscript{D}.
Broad adoption of this method could significantly improve (and sufficiently limit) the interpretation of benchmarking results and highlight the need for reproduced/replicated experimentation.

\section{Discussion}

There are several factors to consider and gaps in existing technology and infrastructure before MIRRER experimentation can be conducted effectively.

\textbf{Context characterization}.
While we define four high-level categories of context parameters that can influence the outcomes of an experiment, we have not specified the manner in which each parameter is recorded.
We aim to strike a balance between ensuring that salient information is recorded and done so in a manner that is digestible, or ``sufficiently complete,'' as the act of recording context should not become too cumbersome and risk discouraging researchers.
There are some existing efforts in the research literature we can leverage to this end (e.g., ASTM standards F3218-19~\cite{astmf4501} for environment characterization and F3381-19~\cite{astmf4503} for describing objects).
More work is needed to develop ``sufficiently complete'' context characterization methods, aiming to outline the unique parameters of context that have impacts on performance and can be controlled during experimentation.

\textbf{Robot system configuration}.
All components of the robot system (especially the CUE) and the interoperability between components must be effectively specified such that it can be reproduced.
On the hardware side, the make and model of robot arms, grippers, and sensors can be recorded, leveraging existing methods like ASTM F3327-23~\cite{astmf4591} for characterizing robot configuration, but characterizing custom hardware will require more details or pointers to documentation.
The use of standard bolt patterns, common mounts, and data connection methods (e.g., ethernet, USB, input/output terminals) simplifies the reproduction of physical connections between components, however publications are not always consistent when reporting robot system configuration~\cite{posecompare}.
On the software side, while enabling technologies like the Robot Operating System (ROS) have become ubiquitous for developing, executing, and connecting components, the field lacks consensus on software structure to ensure compatibility~\cite{posecompare}.
The development of new standards to specify common structures similar to IEEE 1873-2015~\cite{ieee1873} (data formats for navigation maps) can support this issue.

\textbf{Conducting experiments}.
Application of the MIRRER framework relies on researchers' ability to conduct multiple comparable experiments across different labs.
At the very least, this will involve two experiments per lab (iterated experiments) in order to produce a LRR.
While some component-level benchmarking methods are available (e.g., benchmarking motion planners~\cite{liu2022benchmarking}), they largely rely on simulation whereas physical testing requires a full robot system where holistic evaluations are typically conducted.
A replicated experiment assumes that all components other than the CUE are consistent across labs, meaning that no changes are made to the robot system to accommodate the new CUE (e.g., modifying how data is passed between components to accommodate a particular execution pipeline).
A current limitation in the field, though, is the lack of truly modular software components~\cite{posecompare}; the existence of such would not only ease conducting replicated experiments, but would also mitigate the risk of robot system components other than the CUE significantly impacting performance results.
An example of such a software pipeline is GRASPA~\cite{bottarel2023graspa} which has been used to compare the performance of multiple grasp planners.
Such infrastructure as well as the new software standards mentioned previously would also lower the barrier to entry for research labs, both for benchmarking purposes and general utility of open-source software components.

\textbf{Data storage and metrics}.
Effective use of MIRRER throughout the research field requires that results from evaluations be shared and stored in an accessible manner.
Computer vision and machine learning communities regularly utilize assets like the Papers With Code repository~\cite{paperswithcodeweb}, which links benchmarks from contributed papers with submitted scores from users and displays them in a leaderboard.
In those domains, it is much simpler for one to recreate a submitted score using the submitted solution; in most cases, only a computer with the dataset used for benchmarking is required.
As discussed previously, additional guidelines and tools are needed in order to enable this capability in robotics.
Consideration must be given to the size of data that is stored, too, as the difference between reporting a series of performance metrics vs. an entire ROS bag of experiment data is significant.
Similar to the distillation of relevant contextual parameters for characterizing an experiment, the same must be done to determine ``sufficiently complete'' elements that shall be recorded such that measures can be proven and effectively reproduced.
Lastly, the metrics of generalizability and reproducibility are not formally defined and will require additional development.
As a first step, simply reporting the resulting performance metrics from several iterated and reproduced experiments (respectively) and calculating the variance between them can represent either of these metrics.

\textbf{Incentives}.
Despite a general consensus that improving reproducibility and replicability of robotics research is a worthy endeavor, the field does not yet provide sufficient incentives for researchers to do so, such as publication review criteria favoring research that includes comparison benchmarking~\cite{posecompare} (which has become the norm for computer vision research).
The human-robot interaction (HRI) domain does provide opportunities for researchers to publish this type of research by soliciting papers on replication studies at the HRI 2024 conference~\cite{hri2024replication}.
With performance results dependent on a human-in-the-loop, HRI research involves even more variables than what the MIRRER framework is scoped for, but there are no similar venues yet that provide value incentives for reproducing or replicating non-HRI research.

\section{Conclusion and Future Work}

This paper presents a conceptual framework, MIRRER, as a first step towards formalizing how to conduct evaluations of robot systems and their components in terms of generalizability, reproducibility, and replicability.
We intend to continue development of the framework as well as hardware and software tools to enable application of MIRRER for experimentation.
Several experiments have already been conducted within our own lab and with collaborating labs, with more planned now that the initial framework has been defined.





%
%

%
%
%
\bibliographystyle{splncs04}
\bibliography{main.bib}

\end{document}